\pdfoutput=1
\documentclass[11pt]{article}

\usepackage[a4paper,margin=1in]{geometry}
\usepackage[T1]{fontenc}
\usepackage[utf8]{inputenc}

\usepackage{cite}
\usepackage{amsmath,amssymb,amsfonts}
\usepackage{algorithmic}
\usepackage{graphicx}
\usepackage{textcomp}
\usepackage{booktabs}
\usepackage{xcolor}
\usepackage{multirow}
\usepackage{threeparttable}
\usepackage{orcidlink}
\usepackage{indentfirst}

\newcommand{\keywords}[1]{%
  \par\noindent\textbf{Keywords: }#1\par
}

\title{WDFFU-Mamba: A Wavelet-guided Dual-attention Feature Fusion Mamba for Breast Tumor Segmentation in Ultrasound Images}

\author{%
Guoping Cai\orcidlink{0009-0003-4681-8414}
\and Houjin Chen\orcidlink{0000-0002-9247-8495}\thanks{*Corresponding author: Houjin Chen. Postal Address: School of Electronic and Information Engineering, Beijing Jiaotong University, Beijing 100044, China. E-mail: hjchen@bjtu.edu.cn}
\and Yanfeng Li\orcidlink{0000-0002-8441-7721}
\and Jia Sun\orcidlink{0009-0009-0656-6172}
\and Ziwei Chen\orcidlink{0000-0003-0761-6863}
\and Qingzi Geng\orcidlink{0009-0001-7981-8125}%
\thanks{The work was supported in part by the National Natural Science Foundation of China (No. 62272027, 62172029), and the Beijing Natural Science Foundation (No. 4232012, 4244068).}%
\thanks{All authors are with the School of Electronic and Information Engineering, Beijing Jiaotong University, Beijing, China.}%
}

\date{} 

\begin{document}
\maketitle

\begin{abstract}
\textbf{Objective:} Breast ultrasound (BUS) image segmentation plays a vital role in assisting clinical diagnosis and early tumor screening. However, challenges such as speckle noise, imaging artifacts, irregular lesion morphology, and blurred boundaries severely hinder accurate segmentation. To address these challenges, this work aims to design a robust and efficient model capable of automatically segmenting breast tumors in BUS images.
\textbf{Methods:} We propose a novel segmentation network named WDFFU-Mamba, which integrates wavelet-guided enhancement and dual-attention feature fusion within a U-shaped Mamba architecture. A Wavelet-denoised High-Frequency-guided Feature (WHF) module is employed to enhance low-level representations through noise-suppressed high-frequency cues. A Dual Attention Feature Fusion (DAFF) module is also introduced to effectively merge skip-connected and semantic features, improving contextual consistency.
\textbf{Results:} Extensive experiments on two public BUS datasets demonstrate that WDFFU-Mamba achieves superior segmentation accuracy, significantly outperforming existing methods in terms of Dice coefficient and 95th percentile Hausdorff Distance (HD95).
\textbf{Conclusion:} The combination of wavelet-domain enhancement and attention-based fusion greatly improves both the accuracy and robustness of BUS image segmentation, while maintaining computational efficiency.
\textbf{Significance:} The proposed WDFFU-Mamba model not only delivers strong segmentation performance but also exhibits desirable generalization ability across datasets, making it a promising solution for real-world clinical applications in breast tumor ultrasound analysis.
\end{abstract}


\section{Introduction}
\label{sec:introduction}

Breast cancer is one of the most prevalent malignancies among women worldwide. Accurate segmentation of breast tumors is crucial for early screening and the formulation of effective treatment strategies~\cite{ref1}. In recent years, ultrasound imaging has been widely adopted in clinical practice for breast cancer screening and auxiliary diagnosis due to its non-invasive nature, real-time capability, and cost-effectiveness~\cite{ref2}. However, breast ultrasound (BUS) images often suffer from low contrast, strong speckle noise, artifacts, and indistinct lesion boundaries, making automatic segmentation a highly challenging task.

To address these challenges, numerous researchers have turned to deep learning approaches to improve segmentation performance. In particular, convolutional neural network (CNN) architectures based on U-Net~\cite{ref3} have achieved remarkable success in the field of medical image segmentation. Variants such as Attention U-Net~\cite{ref4}, R2U-Net~\cite{ref5}, and UNet++~\cite{ref6} improve feature extraction and semantic fusion capabilities at different levels, and are widely applied to ultrasound image segmentation tasks involving the breast, thyroid, and other organs.

However, CNN is inherently limited in modeling long-range dependencies due to its restricted receptive field, which hinders its ability to capture global contextual information. To overcome this limitation, Transformer-based architectures have been introduced, leveraging self-attention mechanisms to enhance global context modeling. Representative models such as TransUNet~\cite{ref7}, Swin-Unet~\cite{ref8}, and UTNet~\cite{ref9} effectively capture global contextual features and improve the modeling of morphologically complex structures. Chen et al.~\cite{TBME1} proposed a spatial-temporal Transformer framework that combines Swin-Transformer encoding with multi-level feature fusion and global temporal modeling. Meanwhile, recent studies also address practical issues in Transformer-based models. Yue et al. proposed DED-SAM~\cite{ref10}, which distills large vision Transformers for efficient medical segmentation. MedSG-Bench~\cite{ref11} further reveals the limitations of current models in handling sequential medical images, highlighting the need for robust temporal reasoning. Despite these advantages, Transformer architecture suffers from quadratic computational complexity with respect to input size, limiting its suitability for clinical applications that demand real-time processing and computational efficiency.

In recent years, with the advancement of state space model (SSM) theory, researchers have begun exploring its application in medical image segmentation tasks to overcome the limitations of CNN and Transformer in modeling long-range dependencies. Mamba, as an emerging SSM-based architecture, has attracted considerable attention due to its efficiency in long-sequence modeling and linear computational complexity. Ruan et al. first proposed the VM-UNet~\cite{ref12} model, which integrates the Vision Mamba framework into a U-Net architecture to construct an efficient context-aware encoder-decoder structure. The emergence of the Mamba architecture has introduced a new perspective for medical image segmentation research, demonstrating powerful capabilities in handling complex medical images.

For the field of breast tumor segmentation in ultrasound images, many studies have focused on combining local detail extraction with global context modeling to enhance segmentation performance. For example, ESKNet~\cite{ref13} integrates enhanced selective kernel convolution into U-Net, using multi-scale convolution and attention mechanisms to better capture irregular tumor boundaries. MF-Net~\cite{ref14} employs a Transformer-assisted dual encoder and incorporates frequency domain features to improve detail representation. HCTNet~\cite{ref15} fuses CNN and Transformer modules in an interleaved encoder structure, and introduces a spatial-wise cross attention module in the decoder to enhance robustness against complex structures and noise.

Based on the above analysis, we aim to improve the accuracy of tumor segmentation in BUS images from two key aspects: promoting the network's ability in delineating tumor boundaries and enhancing the fusion of semantic information across different stages. To achieve these objectives, we propose a novel Mamba-based network, named WDFFU-Mamba, which integrates wavelet-guided enhancement and dual-attention feature fusion mechanisms. The proposed architecture comprises three core components: a Mamba-based backbone, a Wavelet denoising High-Frequency guided Feature (WHF) module, and a Dual Attention Feature Fusion (DAFF) module.

Specifically, the backbone is built upon a U-shaped Mamba structure that effectively combines local detail preservation with long-range dependency modeling. Within the encoder, the WHF module performs wavelet decomposition on denoised input images to extract high-frequency information. These features are injected into deeper layers via residual connections, thereby enhancing the explicit modeling of boundary regions and improving the precision of tumor contour delineation. At the network's deepest stage, the DAFF module is designed by integrating a lightweight channel attention mechanism and a multi-scale spatial attention mechanism. This dual-attention structure dynamically recalibrates and fuses features across both channel and spatial dimensions, reinforcing the interaction between high-level semantics and low-level spatial details. Comprehensive experiments on two publicly available breast ultrasound image datasets demonstrate that WDFFU-Mamba consistently outperforms mainstream segmentation approaches across multiple metrics. The results highlight its superior accuracy in segmentation and improved sensitivity to boundary structures.

The main contributions of this paper are summarized as follows:
\begin{itemize}
  \item We propose WDFFU-Mamba, a novel segmentation architecture tailored for breast ultrasound image analysis. It achieves an effective trade-off between segmentation accuracy and computational efficiency, while significantly reducing false positives and missed lesion detections commonly encountered in existing methods.
  \item To enhance boundary sensitivity, a Wavelet denoising High-Frequency guided Feature (WHF) module is designed. By performing denoising followed by wavelet decomposition, this module extracts high-frequency components that are injected into the encoder as guidance features, better handling blurry boundaries and structurally complex regions in BUS images.
  \item A Dual Attention Feature Fusion (DAFF) module is constructed based on both channel and spatial attention mechanisms. This module performs dynamic recalibration and fusion of deep and shallow features across multiple dimensions, thereby improving the network's capability in boundary localization and overall segmentation performance.
\end{itemize}

\section{Related Work}
\label{sec:related}

\subsection{Medical Image Segmentation in BUS}
The task of image segmentation is generally categorized into two major types: semantic segmentation and instance segmentation~\cite{ref16,ref17}. Tumor segmentation is a form of semantic segmentation. Early medical image segmentation relied on traditional techniques such as edge detection and region growing. With the advent of deep learning, convolutional neural networks (CNNs) have greatly advanced feature extraction capabilities. U-Net~\cite{ref3}, as a representative CNN model, has been extensively studied and optimized. Variants such as UNet++~\cite{ref6} and SAUNet++~\cite{ref18} further narrow semantic gaps and capture multi-scale features to enhance segmentation accuracy.

However, CNNs are limited in modeling global context, prompting the integration of Transformer-based modules into segmentation networks~\cite{ref19,ref20,ref21,ref22}. Originally developed for NLP, the Transformer architecture~\cite{ref23} has been adapted for vision tasks with models like ViT~\cite{ref24} and Swin Transformer~\cite{ref25}. In medical image segmentation, TransUNet~\cite{ref7} and its derivatives—such as Swin UNETR~\cite{ref26} and TransAttUnet~\cite{ref27}—have demonstrated strong performance by combining CNNs with self-attention.

Recent studies have also highlighted the importance of boundary modeling. Xu et al.~\cite{TBME2} introduced uncertainty-aware consistency constraints to refine boundary regions, while Chen et al.~\cite{TBME3} employed spatial-temporal Transformers to enhance boundary perception in dynamic image sequences.

Specifically for breast ultrasound, advanced architectures have been proposed to extend CNN and Transformer capabilities. FET-UNet~\cite{ref28} integrates ResNet-based CNN branches with Swin Transformer modules in parallel, employing an advanced feature aggregation module for multiscale fusion and improved boundary localization. HAU-Net~\cite{ref29} embeds a Local–Global Transformer block within the skip connections of a U-shaped network and adds a cross-attention mechanism in the decoder to promote inter-layer interaction. BUSSeg~\cite{ref30} further introduces a cross-image dependency module (CDM) using memory banks to model inter-image semantic relationships, enhancing robustness under limited data and noise.

\begin{figure*}[!h]
  \centerline{\includegraphics[width=\textwidth]{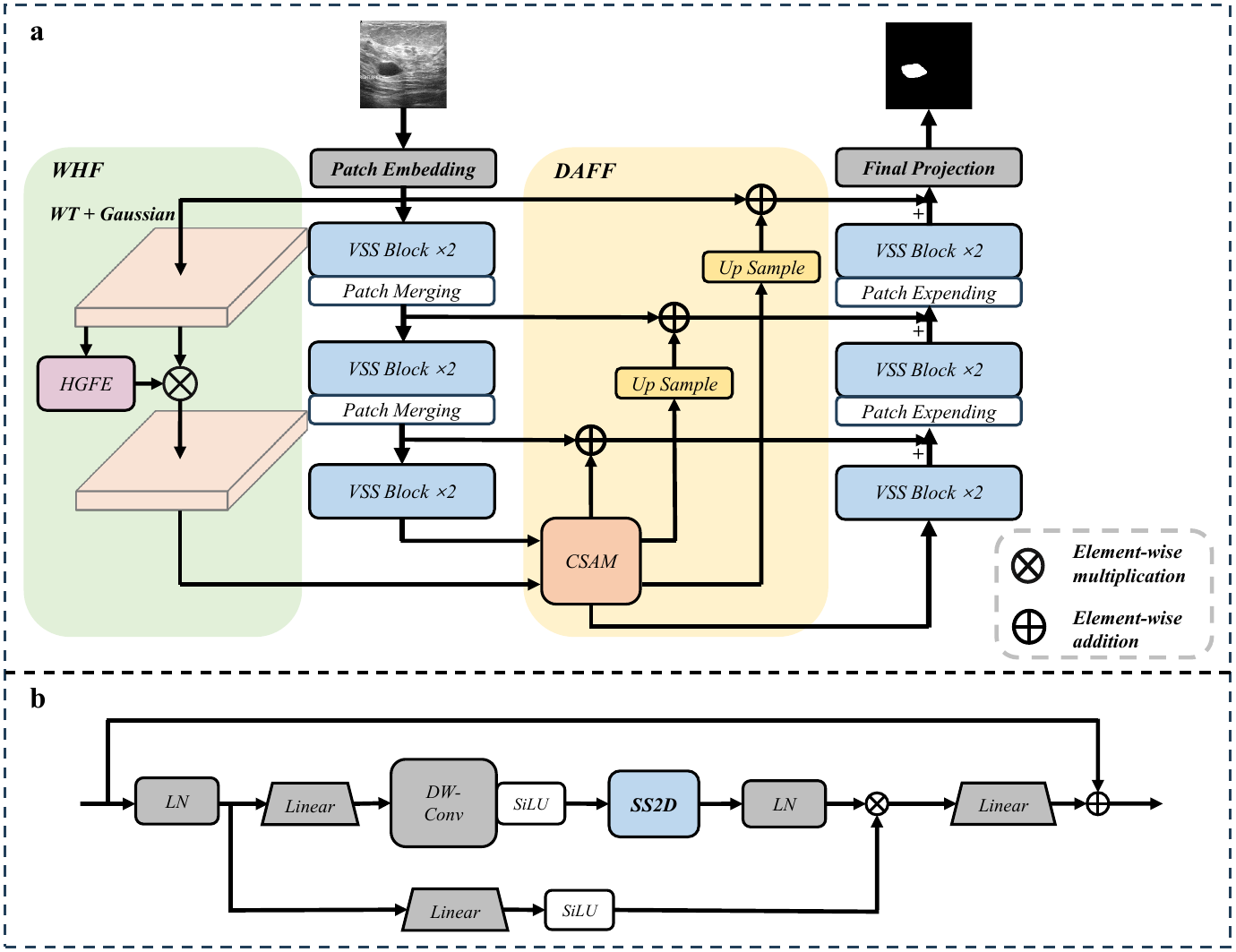}}
  \caption{ (a) Overall architecture of the WDFFU-Mamba network, which consists of a Mamba-based backbone, a WHF module for capturing edge details in feature representations, and a DAFF module that fuses deep and shallow features through dual attention mechanisms; (b) Structure of the VSS Block.}
  \label{fig1}
\end{figure*}

\subsection{Mamba in Medical Image Segmentation}
State-space models (SSMs) have recently shown strong linear-time sequence modeling in NLP. Inspired by this, SSM-based vision models—most notably Mamba~\cite{ref31}—have been applied to segmentation tasks, offering linear complexity versus the quadratic cost of Transformers. Architectures such as ViM~\cite{ref32} and VMamba~\cite{ref33} achieve ViT-comparable performance with reduced complexity.

VM-UNet~\cite{ref12} was the first fully SSM-based segmentation model, employing a pure Vision Mamba backbone within U-Net and setting a benchmark on multiple datasets. Building on this, VM-UNetV2~\cite{ref34} improves representational capacity by initializing with pretrained VMamba weights and adding multi-scale deep supervision, yielding significant gains.

Subsequent work has refined Mamba-based designs to capture spatial context and enhance robustness. HC-Mamba~\cite{ref35} incorporates dilated and depthwise separable convolutions into the Mamba module to balance receptive fields and efficiency. H-VMUNet~\cite{ref36} introduces a High-Order 2D Selective Scanning module and a Local Feature Learning module to reduce redundant features while strengthening local detail. MSVM-UNet~\cite{ref37} employs multi-scale convolution and large-kernel upsampling to address directional sensitivity in 2D segmentation.

To tackle complexity and data scarcity, lightweight and semi-supervised variants have emerged. LightMUNet~\cite{ref38} streamlines U-Net with Mamba for lower parameter counts. Semi-MambaUNet~\cite{ref39} uses a dual-branch network for pseudo-labeling and pixel-level self-supervised contrastive learning. U-Mamba~\cite{ref40} proposes a general CNN–SSM hybrid, and Swin-U-Mamba~\cite{ref41} demonstrates the benefits of ImageNet pretraining on Mamba-based decoders.

\subsection{Attention Mechanisms}
Attention mechanisms have been widely adopted to capture contextual dependencies. Attention U-Net~\cite{ref4} integrates attention gates into skip connections to suppress background and highlight foreground features. ATFE-Net~\cite{ref42} introduces an axial Transformer module to model long-range dependencies efficiently along spatial axes. CFATransUNet~\cite{ref43} designs a Channel-wise Cross-Fusion Transformer and Attention module for cross-layer semantic fusion and channel weighting, reducing redundancy. MF-Net~\cite{ref14} adds an auxiliary Transformer encoder and a Global Feature Enhancement module to fuse multi-source semantics, improving sensitivity to lesion regions. However, most existing attention modules focus on either low-level or high-level features and often incur high computational costs, especially on high-resolution medical images.

\section{Methodology}
\label{sec:method}

\subsection{Overview}
The overall architecture of the proposed WDFFU-Mamba model is illustrated in Fig.~\ref{fig1}(a). This network builds upon the three-level VM-UNet~\cite{ref12} framework, which utilizes the Visual State Space (VSS) block as its core component. The structure of the VSS block is depicted in Fig.~\ref{fig1}(b). Within the VSS block, the 2D Selective Scan (SS2D) operation unfolds the input image into sequences along four distinct directions: top-left to bottom-right, bottom-right to top-left, top-right to bottom-left, and bottom-left to top-right. These sequences are then processed by the S6 block, which extracts features and effectively captures multi-directional information from the image.

To enhance the encoder's capability, we introduce two modules: the Wavelet denoising High-Frequency guided Feature (WHF) module and the Dual Attention Feature Fusion (DAFF) module, inserted at the deepest layer of the network. The WHF module addresses the prevalent issue of high noise in ultrasound images and enhances the network's ability to retain edge information. It comprises a wavelet denoising unit and a high-frequency feature extraction unit. The former effectively suppresses noise, while the latter extracts critical high-frequency edge information from the denoised feature maps. This high-frequency information is then transmitted to deeper layers through residual connections, improving the representation of edge details. The DAFF module facilitates the deep fusion of features from the encoder and the WHF module. The DAFF module incorporates redesigned spatial and channel attention mechanisms, enabling separate modeling of the two input features. It generates spatial and channel attention maps to weight the summed input features, thereby producing the final fused map. This fused feature map is then adjusted in channel numbers and upsampled before passing through skip connections to each decoder level, enhancing the feature reconstruction capability during the decoding phase.

\begin{figure}[!t]
  \centering
  \includegraphics[width=\textwidth]{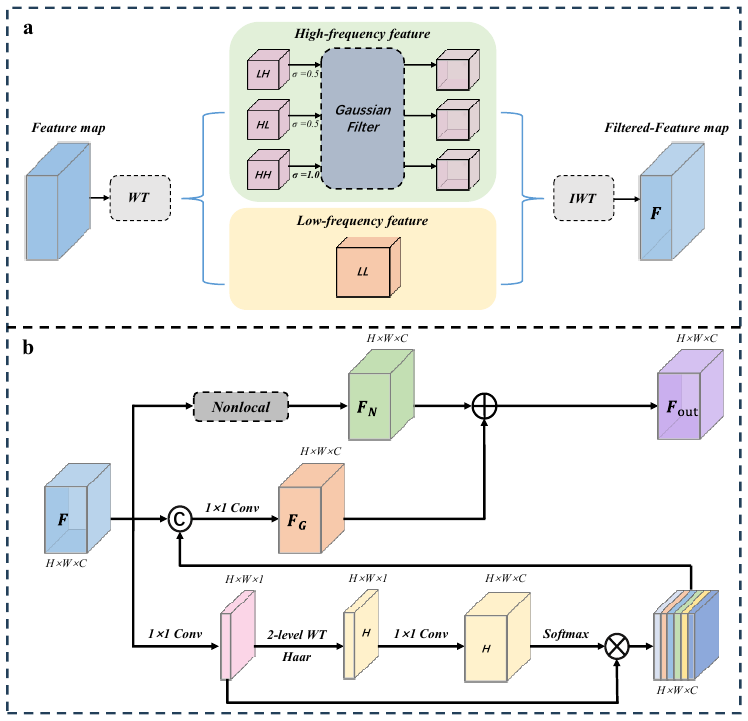} 
  \caption{(a) Wavelet denoising process in the WHF module, including wavelet transform, Gaussian filtering, and inverse wavelet transform; (b) High-Frequency Guided Feature Extraction (HGFE) in the WHF module.}
  \label{fig2}
\end{figure}

\subsection{WHF}
Speckle noise is a common and complex physical artifact in ultrasound imaging, typically characterized by random intensity fluctuations. This type of noise poses significant challenges to tumor segmentation. To mitigate its impact, a Wavelet denoising High-Frequency guided Feature (WHF) module is designed. The detailed process is illustrated in Fig.~\ref{fig2}(a).

Wavelet denoising is a multi-scale signal processing technique based on wavelet transform, which is capable of effectively removing noise while preserving important structural features in the image. The wavelet transform decomposes an image into components at multiple frequency scales, including low-frequency components that contain the main structural information and high-frequency components that typically include both image details and noise. In this study, a single-level Haar wavelet is employed to decompose the feature map \(X\) into one low-frequency component (LL) and three high-frequency components: the horizontal high-frequency component (LH), vertical high-frequency component (HL), and diagonal high-frequency component (HH). Among these, the HH component is particularly sensitive to fine textures and speckle noise. Considering the random and localized characteristic of speckle noise, we apply a stronger Gaussian filter specifically to the HH component to suppress noise interference. After filtering, the denoised high-frequency components and the original low-frequency component are recombined using inverse wavelet transform, resulting in a reconstructed feature map \(F\). This map preserves crucial edge information while significantly reducing the influence of high-frequency noise.

The wavelet denoising process can be formally described as follows:
\begin{gather}
\Delta\bigl(F_{\mathrm{original}}\bigr)
  = (LL,\;LH,\;HL,\;HH)
  \label{eq:whf_decomp}\\
\begin{aligned}
  \widehat{LH} &= G_{\sigma=0.5}(LH),\\
  \widehat{HL} &= G_{\sigma=0.5}(HL),\\
  \widehat{HH} &= G_{\sigma=1.0}(HH)
\end{aligned}
  \label{eq:whf_filter}\\
F = \Delta^{-1}\bigl(LL,\;\widehat{LH},\;\widehat{HL},\;\widehat{HH}\bigr)
  \label{eq:whf_recon}
\end{gather}

where $\Delta$ and $\Delta^{-1}$ denote the single‐level Haar wavelet transform and its inverse transform. The size of $F_{\mathrm{original}}$ and $F$ are $H\times W\times C$. The size of $LL$, $LH$, $HL$ and $HH$ are $\tfrac H4\times\tfrac W4\times C$.

The reconstructed feature map is subsequently fed into the High‐Frequency Guided Feature Extraction (HGFE) module to further enhance high‐frequency edge features. The structure of this module is illustrated in Fig.~\ref{fig2}(b). HGFE consists of two sub‐branches. Branch 1 is the High‐Frequency Guidance Branch. The input feature map is first processed with a convolutional operation, followed by a two‐level Haar wavelet decomposition to extract more refined high‐frequency feature maps. Next, a Softmax function is applied to compute channel‐wise weights for the high‐frequency components, generating a high‐frequency attention map. This attention map is element‐wise multiplied with the single‐channel representation of the input feature map to produce the high‐frequency enhanced feature map. Finally, this feature map is concatenated with the original input along the channel dimension to form the fused feature map $F_1$. This process for Branch 1 can be formally described as follows:
\begin{gather}
F_{1} = \mathrm{Conv}(F) \label{eq:4}\\
F_{G} = \mathrm{Conv}\Bigl(\mathrm{Cat}\bigl(\Theta\bigl(\mathrm{Conv}(\Delta^{2}(F_{1}))\bigr)\otimes F_{1},\;F\bigr)\Bigr)
\label{eq:5}
\end{gather}
where \(\mathrm{Conv}(\cdot)\) denotes the \(1\times1\) convolution, which is used to adjust the number of output channels and \(F_{1}\in\mathbb{R}^{1\times H\times W}\); \(\mathrm{Cat}(\cdot,\cdot)\) denotes the concatenation of feature maps; \(\Theta(\cdot)\) represents the Softmax function.

Branch 2 feeds the input feature map \(F\) into a Non-local module to capture global contextual information, resulting in a non-local feature map \(F_{N} \in \mathbb{R}^{C \times H \times W}\). Subsequently, we pass \(F_{G}\) through a channel adjustment layer and add it with \(F_{N}\), yielding the final enhanced feature map. This process for Branch 2 can be formally described as follows:
\begin{equation}
\begin{gathered}
F_{N} = \mathrm{Nonlocal}(F) \\
F_{\mathrm{out}} = F_{N} \oplus F_{G}
\end{gathered}
\label{eq:6}
\end{equation}

\begin{figure}[!h]
  \centerline{\includegraphics[width=\columnwidth]{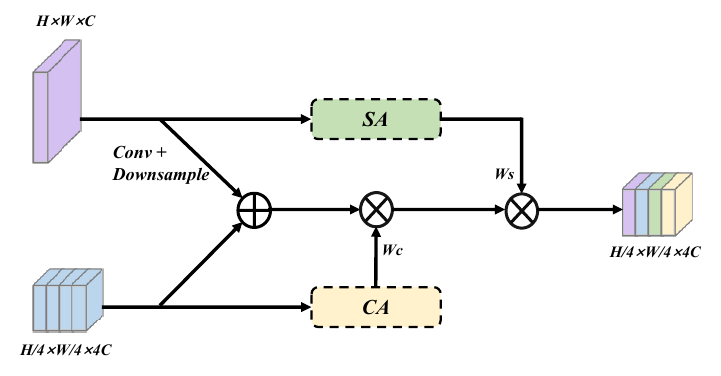}}
  \caption{Structure of the DAFF module. Two different inputs are processed through spatial and channel attention modules, respectively, and the resulting attention weights are multiplied with the fused features.}
  \label{fig3}
\end{figure}

\subsection{DAFF}

In the encoder part of the network, a common design involves progressively reducing the spatial resolution of feature maps and increasing the number of channels as the network depth increases, in order to more effectively learn semantic representations. However, while this downsampling process enhances high-level semantic features, it inevitably leads to a significant reduction in the spatial resolution of deep feature maps, resulting in the loss of critical details—particularly along tumor boundaries. To mitigate this issue, inspired by the concept of residual connections~\cite{ref44}, the high-frequency guided features extracted by the WHF module, as introduced in the previous section, are incorporated into the deeper layers of the network via residual connections. This design enhances the decoder's capability to recover edge and texture information during the reconstruction process.

Nevertheless, due to inherent differences in feature dimensions and distributions between shallow high-resolution features and deep semantic features, directly fusing them may introduce additional noise and degrade the overall performance. To address this challenge, we design a Dual Attention Feature Fusion (DAFF) module, which applies both spatial attention and channel attention mechanisms to adaptively reweight the importance of different input features. The fused features are then element-wise multiplied by their corresponding attention weights, enhancing the multi-level feature integration, suppressing noise interference, and improving both representation capacity and segmentation performance.

The overall structure of the DAFF module is illustrated in Fig.~\ref{fig3}. The module takes two inputs: high-resolution features (HF, with fewer channels) and low-resolution features (LF, with more channels). The LF first passes through a submodule based on channel attention. This submodule draws on the classical design of channel attention mechanisms and introduces an optimization by replacing fully connected layers with \(1\times1\) convolutions, achieving higher computational efficiency and reduced parameter overhead. This replacement not only lowers the model complexity but also retains the ability to effectively model channel-wise dependencies. The final output of this submodule is a channel attention weight map \(M_c\). The computation process of \(M_c\) is as follows:
\begin{equation}
\begin{gathered}
z_{\mathrm{avg}} = \mathrm{GAP}(LF) = \frac{1}{H \times W} \sum_{i=1}^{H} \sum_{j=1}^{W} LF_{c,ij} \\
z_{\mathrm{max}} = \mathrm{GMP}(LF) = \max_{i,j} LF_{c,ij}
\end{gathered}
\label{eq:7}
\end{equation}
\begin{equation}
M_{c} = \sigma\left( \mathrm{Conv}_{1\times1}(z_{\mathrm{avg}}) + \mathrm{Conv}_{1\times1}(z_{\mathrm{max}}) \right)
\label{eq:8}
\end{equation}
where \( LF \in \mathbb{R}^{4C \times \frac{H}{4} \times \frac{W}{4}} \), \( z_{\mathrm{avg}}, z_{\mathrm{max}} \in \mathbb{R}^{4C \times 1 \times 1} \), and \( \sigma(\cdot) \) represents the Sigmoid function.

For the HF features, we design a novel spatial attention (SA) submodule. Unlike conventional spatial attention mechanisms, this module employs Wavelet Transform Convolution (WTC) in place of standard convolution operations. Wavelet transform is capable of capturing features in both the time and frequency domains. When combined with convolutional operations, it not only facilitates multi-scale feature extraction but also enhances the representation of edges and fine details. As a result, WTC improves the spatial attention module's sensitivity to local information. The output of this submodule is denoted as \(M_s\). The workflow of the SA module can be described as follows:
\begin{gather}
F_{\mathrm{avg}} = \mathrm{MeanPool}(\mathrm{HF}), \quad 
F_{\mathrm{max}} = \mathrm{MaxPool}(\mathrm{HF})
\label{eq:9}\\
M_s = \sigma\left(\mathrm{WTC}_{7\times7}\left(\mathrm{cat}(F_{\mathrm{avg}}, F_{\mathrm{max}})\right)\right)
\label{eq:10}
\end{gather}
where \( \mathrm{HF} \in \mathbb{R}^{C \times H \times W} \), 
\( F_{\mathrm{avg}}, F_{\mathrm{max}} \in \mathbb{R}^{1 \times H \times W} \), and 
\( \mathrm{WTC}_{7\times7}(\cdot) \) represents a wavelet convolution with a kernel size of 7.

Finally, the fused feature map is element-wise multiplied by the spatial attention weight \( M_s \) and the channel attention weight \( M_c \), resulting in an enhanced feature \( F_{\mathrm{DAFF}} \). This enhanced feature not only preserves essential semantic information but also further strengthens the representation of edge and structural details. After performing upsampling and channel adjustment, the enhanced feature is fed into each stage of the decoder as the input for skip connections, thereby participating in the feature reconstruction and restoration process.
\begin{equation}
F_{\mathrm{out}} = \left( \mathrm{Conv}(\mathrm{Maxpool}(\mathrm{HF})) \oplus \mathrm{LF} \right) \odot M_c \odot M_s
\label{eq:11}
\end{equation}

\subsection{Loss Function}

In this study, we employ a weighted combination of Binary Cross-Entropy Loss (BCE) and Dice Loss as the loss function, which jointly accounts for prediction accuracy and class imbalance in the image segmentation task. The Binary Cross-Entropy Loss measures the discrepancy between the predicted values and the ground truth in binary classification settings. It is defined as follows:
\begin{equation}
L_{\mathrm{BCE}} = -\frac{1}{N} \sum_{i=1}^{N} \left[ y_i \cdot \log(\hat{y}_i) + (1 - y_i) \cdot \log(1 - \hat{y}_i) \right]
\label{eq:12}
\end{equation}
where \( y_i \) denotes the ground truth label, \( \hat{y}_i \) represents the predicted value, and \( N \) is the number of samples.

Dice Loss is used to measure the area overlap between the predicted results and the ground truth labels. It demonstrates stronger robustness to class imbalance, particularly in image segmentation tasks. The definition is as follows:
\begin{equation}
L_{\mathrm{Dice}} = 1 - \frac{2 \sum_{i=1}^{N} y_i \hat{y}_i + \epsilon}
{\sum_{i=1}^{N} y_i^2 + \sum_{i=1}^{N} \hat{y}_i^2 + \epsilon}
\label{eq:13}
\end{equation}

The final loss function is a combination BCE Loss and Dice Loss, expressed as follows. It demonstrates strong overall performance in image segmentation tasks, effectively balancing classification accuracy and structural overlap.
\begin{equation}
L = 0.5 L_{\mathrm{BCE}} + L_{\mathrm{Dice}}
\label{eq:14}
\end{equation}

\section{Experiments}

\subsection{Datasets}

In this study, two publicly available breast ultrasound image datasets are used to evaluate the performance of the proposed method. The first dataset, BUSI, was constructed by Al-Dhabyani et al. in 2020~\cite{ref45}. It contains 780 images with an average resolution of 500×500 pixels. Images are categorized into three classes: normal, benign, and malignant. To specifically focus on the segmentation of breast lesions, only the images labeled as benign and malignant are used in this study, while normal breast images are excluded. 

The second dataset, BUS, was collected by Yap et al.~\cite{ref46}, comprising 163 carefully annotated images with an average resolution of 760×570 pixels. Among them, 53 images depict malignant tumors and 110 depict benign tumors. For both datasets, the benign and malignant images are split into training, validation, and test sets in a 70\%:15\%:15\% ratio. The model weights that achieve the best performance on the validation set are saved, and the final segmentation performance is evaluated on the test set.

\subsection{Evaluation Metrics}

In this study, six evaluation metrics are used to quantitatively evaluate the tumor segmentation performance of the model on breast ultrasound images, including the Dice coefficient, 95\% Hausdorff Distance (HD95), Jaccard index, precision, recall, and sensitivity. Their corresponding definitions are listed below:
\begin{equation}
\text{Dice} = \frac{2TP}{2TP + FP + FN}
\tag{15}
\end{equation}

\begin{equation}
\text{Jaccard} = \frac{TP}{TP + FP + FN}
\tag{16}
\end{equation}

\begin{equation}
\text{Precision} = \frac{TP}{TP + FP}
\tag{17}
\end{equation}

\begin{equation}
\text{Recall} = \frac{TP}{TP + FN}
\tag{18}
\end{equation}

\begin{equation}
\text{Specificity} = \frac{TN}{TN + FP}
\tag{19}
\end{equation}
Here, $TP$ and $TN$ represent the number of correctly segmented breast lesion pixels and background pixels, respectively. $FP$ denotes the number of background pixels incorrectly segmented as lesion pixels, while $FN$ refers to lesion pixels incorrectly predicted as background. For the five metrics mentioned above, values closer to 1 indicate better segmentation performance.

The 95\% Hausdorff Distance (HD95) is a robust metric for evaluating the boundary alignment between the predicted segmentation and the ground truth. Unlike the traditional Hausdorff Distance, which measures the maximum mismatch between two point sets, HD95 computes the 95th percentile of all minimum distances between boundary points of the two regions. This helps eliminate outlier-induced deviations and provides a more stable and representative measure of segmentation precision. A lower HD95 value indicates better alignment and segmentation performance.
\begin{equation}
\begin{aligned}
H(A, B) &= \max \left( \max_{a \in A} \min_{b \in B} \| a - b \|,\ \max_{b \in B} \min_{a \in A} \| b - a \| \right) \\
H_{95\%}(A, B) &= 95\%\text{-quantile} \Big( \min_{b \in B} \| a - b \|,\ \\
&\qquad\qquad\qquad\qquad \text{for each } a \in A \Big)
\end{aligned}
\tag{20}
\end{equation}
where $A$ and $B$ represent the boundary point sets of the predicted and ground truth regions, respectively, and $\|\cdot\|$ denotes the Euclidean distance.

\subsection{Implementation Details}
In our experiments, all images—including those used for training, validation, and testing—were resized to $224\times224$ pixels. To prevent overfitting, data augmentation techniques were applied, including random flipping and random rotation. During training, we adopted the AdamW optimizer to update the network parameters, with an initial learning rate set to $1\times10^{-3}$ and a weight decay of $1\times10^{-2}$.

Consistent with the VM-UNet study, we initialized our network using the pretrained weights of VMamba-S~\cite{ref36}, which was trained on the ImageNet-1k dataset. The model was trained for 300 epochs with a batch size of 12. The development environment consisted of Ubuntu 18.04, Python 3.8.19, and PyTorch 1.13. All experiments were conducted on a single NVIDIA GeForce GTX 3090 GPU.

\subsection{Ablation Studies}
In order to validate the effectiveness of each proposed component, we conduct a series of ablation experiments on two datasets. The experiments are based on a U-shaped Vision Mamba architecture, specifically adopting a three-layer encoder--decoder variant VM-UNet~\cite{ref12}. In this configuration, the number of VSS Blocks in each layer of the encoder is set to [2, 2, 2], and the decoder is configured similarly with [2, 2, 2]. To ensure fairness and comparability, all ablation experiments are conducted under the same hardware and software environment.

\begin{figure}[!h]
  \centering
  \includegraphics[width=0.5\textwidth]{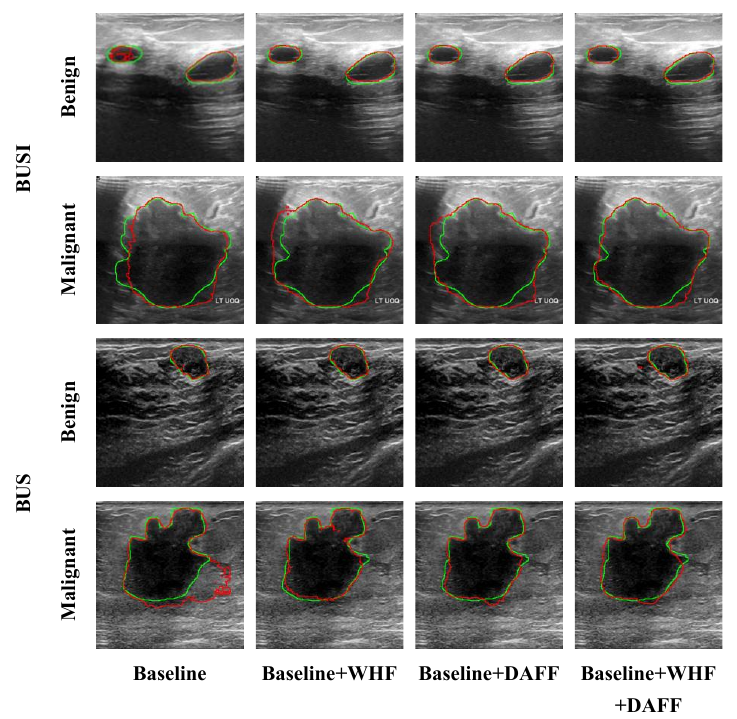}
  \caption{Visualization results of the ablation study. The labels on the left indicate the dataset (including BUSI and BUS) and the tumor type (benign or malignant), while the labels below indicate different combinations of network components. Green contours represent the ground truth, and red contours represent the predicted results.}
  \label{fig4}
\end{figure}

\begin{figure}[!h]
  \centering
  \includegraphics[width=0.5\textwidth]{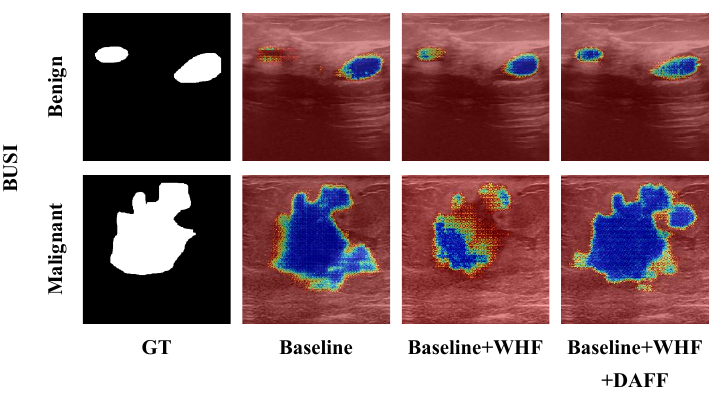}
  \caption{Attention heatmaps of the network. The blue regions indicate the areas of primary focus by the network.}
  \label{fig5}
\end{figure}

Based on this baseline model, we develop three improved variants: Baseline + WHF, Baseline + DAFF, and Baseline + WHF + DAFF. All models are trained and evaluated on the BUS and BUSI ultrasound image datasets. Experimental results are presented in Table~\ref{tab:ablation}, while the visualized outcomes of the ablation studies are shown in Fig.~\ref{fig4} and Fig.~\ref{fig5}.

\begin{table*}[!h]
\centering
\begin{threeparttable}
\caption{Ablation study of different network components.}
\label{tab:ablation}
\renewcommand{\arraystretch}{1.2}
\begin{tabular}{lcccc}
\hline
\textbf{Method} & \multicolumn{2}{c}{\textbf{BUSI}} & \multicolumn{2}{c}{\textbf{BUS}} \\
& \textbf{Dice (\%)} & \textbf{HD95} & \textbf{Dice (\%)} & \textbf{HD95} \\
\hline
Baseline & 81.42$\pm$2.27 & 26.78$\pm$5.70 & 82.12$\pm$3.79 & 19.57$\pm$8.27 \\
Baseline + WHF & {\color{blue}83.10$\pm$3.01} & 18.93$\pm$6.75 & {\color{blue}84.92$\pm$1.91} & {\color{blue}22.22$\pm$5.72} \\
Baseline + DAFF & 82.41$\pm$3.93 & {\color{blue}16.08$\pm$6.73} & 82.62$\pm$2.02 & 22.44$\pm$4.97 \\
Baseline + WHF + DAFF & {\color{red}\textbf{84.48$\pm$2.89}} & {\color{red}\textbf{13.28$\pm$5.17}} & {\color{red}\textbf{85.20$\pm$1.75}} & {\color{red}\textbf{17.90$\pm$4.38}} \\
\hline
\end{tabular}
\vspace{2pt}
\begin{tablenotes}
\setlength{\leftskip}{0pt}
\small
\item The best results are highlighted in red and the second-best in blue. 
\item All evaluation metrics are reported as mean ± standard deviation.
\end{tablenotes}
\end{threeparttable}
\end{table*}

According to the experimental results, the Baseline + WHF method improves the Dice coefficient by 3.5\% and reduces the HD95 by 4.56 on the BUSI dataset. On the BUS dataset, the Dice score increases by 0.98\% and the HD95 decreases by 0.64. These results indicate that the WHF module, by introducing high-frequency guided features, effectively enhances the edge representation capability in deeper layers of the network, thereby improving the accuracy of the decoding stage. On the other hand, the Baseline + DAFF method achieves a 1.2\% increase in Dice and a 4.34 reduction in HD95 on the BUSI dataset compared to the baseline model. This demonstrates that the Dual Attention Feature Fusion (DAFF) module performs well in integrating shallow and deep features, leading to improved feature representation and target segmentation.

Overall, the Baseline + WHF + DAFF method outperforms both Baseline + WHF and Baseline + DAFF across the BUSI and BUS datasets, suggesting that the strategy of enhancing shallow features and effectively integrating them with deep semantic features via skip connections contributes to further improvements in segmentation performance.

In addition, the visualized segmentation results in Fig.~\ref{fig4} intuitively show the performance of each model across different lesion types (benign and malignant). The comparison between green (ground truth) and red (predicted) contours reveals that the Baseline + WHF + DAFF model achieves the most accurate boundary fitting, especially in malignant cases from the BUSI dataset, where it better captures complex tumor edges with reduced error. The attention heatmaps in Fig.~\ref{fig5} further validate these observations. It can be clearly seen that with the progressive inclusion of the WHF and DAFF modules enhances the model's attention mechanism, leading to a more precise focus on target regions. This suggests that the proposed modules effectively enhance both feature discriminability and spatial focus, promoting better recognition of critical regions.

Based on the above analysis, we conclude that the proposed WDFFU-Mamba network significantly improves breast ultrasound image segmentation performance, particularly in boundary restoration and attention localization, demonstrating strong potential for clinical application in computer-aided diagnosis.

\subsection{Comparison Experiments}

To further demonstrate the superior segmentation performance of our proposed network, we conduct comparative experiments against eight state-of-the-art image segmentation models. These models include the classical U-Net~\cite{ref3} and its two variants (Attention U-Net~\cite{ref4} and UNet++~\cite{ref6}), the representative hybrid Transformer--CNN model TransUNet~\cite{ref7}, the pure Transformer-based network MISSFormer~\cite{ref47}, two recent Mamba-based segmentation networks, namely H-vmunet~\cite{ref36} and UltraLight-VM-UNet~\cite{ref48}, as well as the full version of our baseline model, VM-UNet~\cite{ref12}.

\begin{figure*}[!h]
  \centering
  \includegraphics[width=\textwidth]{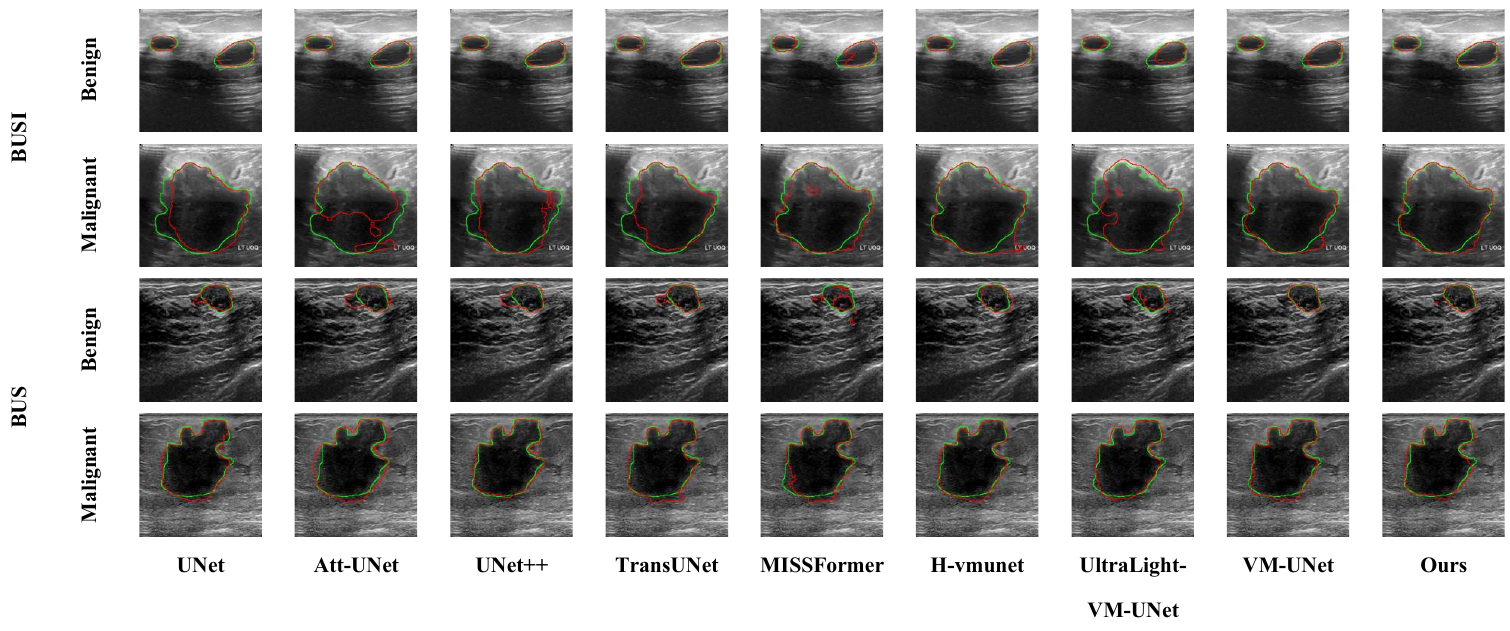}
  \caption{Visualization results of the comparative experiments. The labels below indicate eight state-of-the-art medical image segmentation methods and the proposed method.}
  \label{fig6}
\end{figure*}

\begin{table}[!h]
\centering
\caption{Comparison of segmentation results on the BUSI and BUS datasets across different methods.}
\label{tab2}
\renewcommand{\arraystretch}{1.2}

\resizebox{\textwidth}{!}{%
\begin{tabular}{l l c c c c c c c}
\hline
\textbf{Dataset} & \textbf{Method} & \textbf{Dice (\%)} & \textbf{HD95} & \textbf{Jaccard (\%)} & \textbf{Precision (\%)} & \textbf{Recall (\%)} & \textbf{Specificity (\%)} & $p$-value \\
\hline
\multirow{9}{*}{BUSI}
& UNet & 79.40$\pm$2.25 & 53.51$\pm$8.58 & 70.22$\pm$2.47 & 82.47$\pm$2.38 & 83.31$\pm$2.07 & {\color{blue}98.11$\pm$0.40} & $<$0.05 \\
& Att-UNet & 77.45$\pm$2.37 & 52.87$\pm$8.32 & 67.91$\pm$2.57 & 79.09$\pm$2.40 & {\color{blue}84.67$\pm$2.29} & 97.63$\pm$0.47 & $<$0.001 \\
& UNet++ & 80.19$\pm$2.09 & 46.29$\pm$7.88 & 70.89$\pm$2.38 & 85.02$\pm$2.17 & 82.62$\pm$2.07 & {\color{red}\textbf{98.60$\pm$0.26}} & $<$0.05 \\
& TransUNet & {\color{blue}82.53$\pm$2.00} & {\color{blue}19.95$\pm$3.75} & {\color{blue}73.78$\pm$2.19} & {\color{red}\textbf{89.18$\pm$1.76}} & 80.35$\pm$2.28 & 97.23$\pm$1.44 & $<$0.05 \\
& MISSFormer & 76.04$\pm$2.36 & 42.28$\pm$4.96 & 65.89$\pm$2.51 & 79.41$\pm$2.44 & 78.63$\pm$2.43 & 97.21$\pm$1.09 & $<$0.001 \\
& H-vmunet & 79.57$\pm$2.07 & 43.36$\pm$6.28 & 69.71$\pm$2.24 & 76.31$\pm$2.33 & {\color{red}\textbf{87.93$\pm$1.79}} & 97.37$\pm$0.42 & $<$0.05 \\
& UltraLight-VM-UNet & 73.97$\pm$2.62 & 38.85$\pm$6.69 & 63.95$\pm$2.67 & 77.14$\pm$2.73 & 80.10$\pm$2.50 & 95.64$\pm$1.47 & $<$0.001 \\
& VM-UNet & 81.13$\pm$2.30 & 22.95$\pm$5.07 & 72.69$\pm$2.42 & {\color{blue}88.78$\pm$2.13} & 78.61$\pm$2.34 & 96.86$\pm$1.47 & $<$0.05 \\
& \textbf{Ours} & {\color{red}\textbf{85.20$\pm$1.75}} & {\color{red}\textbf{17.90$\pm$4.38}} & {\color{red}\textbf{77.01$\pm$1.95}} & 88.70$\pm$1.80 & 84.54$\pm$1.94 & 97.95 & -- \\
\hline
\multirow{9}{*}{BUS}
& UNet & 68.63$\pm$6.14 & 40.99$\pm$13.21 & 58.92$\pm$5.77 & 77.56$\pm$6.06 & 68.81$\pm$6.56 & 91.64$\pm$5.29 & $<$0.05 \\
& Att-UNet & 65.74$\pm$6.21 & 36.67$\pm$9.44 & 55.67$\pm$5.88 & 68.24$\pm$6.40 & 73.77$\pm$6.75 & 91.09$\pm$5.26 & $<$0.05 \\
& UNet++ & 78.20$\pm$4.12 & 43.25$\pm$13.51 & 68.01$\pm$4.57 & 82.38$\pm$3.92 & 82.40$\pm$4.26 & {\color{red}\textbf{99.37$\pm$0.13}} & $<$0.05 \\
& TransUNet & {\color{blue}79.45$\pm$4.55} & 29.17$\pm$12.74 & {\color{blue}70.41$\pm$4.83} & 79.06$\pm$4.81 & {\color{blue}85.38$\pm$4.69} & 99.05$\pm$0.27 & $<$0.05 \\
& MISSFormer & 70.96$\pm$4.42 & 45.21$\pm$11.81 & 59.01$\pm$4.76 & 76.78$\pm$4.34 & 77.97$\pm$5.11 & {\color{blue}99.32$\pm$0.13} & $<$0.05 \\
& H-vmunet & 75.10$\pm$4.57 & 24.20$\pm$7.04 & 64.29$\pm$4.63 & 76.87$\pm$4.23 & 81.25$\pm$5.53 & 95.3$\pm$3.81 & $<$0.05 \\
& UltraLight-VM-UNet & 71.80$\pm$3.06 & 39.83$\pm$9.48 & 58.23$\pm$3.75 & 71.09$\pm$4.22 & 80.53$\pm$4.13 & 98.83$\pm$0.31 & $<$0.001 \\
& VM-UNet & 79.04$\pm$5.36 & {\color{red}\textbf{7.81$\pm$3.23}} & 71.21$\pm$5.37 & {\color{red}\textbf{83.14$\pm$5.20}} & 79.26$\pm$5.86 & 91.95$\pm$5.30 & $<$0.05 \\
& \textbf{Ours} & {\color{red}\textbf{84.48$\pm$2.89}} & {\color{blue}13.28$\pm$5.17} & {\color{red}\textbf{75.42$\pm$3.69}} & {\color{blue}82.71$\pm$3.38} & {\color{red}\textbf{90.31$\pm$3.04}} & 99.24 & -- \\
\hline
\end{tabular}%
} 

\vspace{2pt}
\begin{minipage}{\textwidth}
\small
\textit{Notes.} The best results are highlighted in red and the second-best results in blue.
All evaluation metrics are reported as mean $\pm$ standard deviation.
$p$-values are calculated via $t$-tests on Dice scores between each method and ours.
\end{minipage}

\end{table}

To ensure fairness, all comparison experiments are conducted under the same computational environment. The visual segmentation results of each method are shown in Fig.~\ref{fig6}, and the corresponding quantitative evaluation results are summarized in Table~\ref{tab2}.

As shown in Table~\ref{tab2}, the proposed method demonstrates significant performance advantages or strong competitiveness across all six mainstream evaluation metrics, with particularly outstanding results in the two key indicators: Dice coefficient and HD95. The Dice coefficient measures the overlap between the predicted segmentation and the ground truth, serving as a core metric for evaluating segmentation accuracy. In contrast, HD95 reflects the maximum deviation between the predicted and actual boundaries, offering a more sensitive assessment of the model’s precision and robustness in boundary handling. Specifically, on the BUSI dataset, the proposed method improves the Dice coefficient by 2.68\% and reduces HD95 by 2.05 compared to the second-best method. On the BUS dataset, despite a slightly lower performance in Specificity, our method achieves the best or second-best results in the remaining five metrics. Notably, the Dice coefficient is improved by 5.03\% over the second-best method.

These results clearly demonstrate that the proposed method achieves high-precision segmentation on both BUS and BUSI datasets. In particular, it shows remarkable advantages in terms of region overlap and boundary localization, validating the effectiveness and practicality of the proposed WHF and DAFF modules in enhancing medical image segmentation performance.

\subsection*{4.6 Generalization Experiments}
To further evaluate the generalization capability of the model, we adopt a cross-dataset testing strategy, in which the model is trained on the BUSI dataset and directly evaluated on the test set of the BUS dataset. For the model trained on BUSI, the BUS dataset remains entirely unseen. The experimental results are presented in Table~\ref{tab3}.

As shown in Table~\ref{tab3}, although our method only has a parameter count of 5.65M, which is significantly lower than that of VM-UNet, it still achieves comparable segmentation performance in the cross-dataset testing scenario (Dice: 75.70 vs. 75.16, Jaccard: 65.06 vs. 66.84). These results indicate that our method maintains strong generalization ability and superior parameter efficiency, even with a substantially reduced model size.

\begin{table}[!h]
\centering
\caption{Comparison of parameter statistics and generalization segmentation results across different methods.}
\label{tab3}

\vspace{0.5em}

\resizebox{\textwidth}{!}{%
\begin{tabular}{@{}lccccccc@{}}
\hline
\textbf{Method} & \textbf{Params} & \textbf{Dice(\%)} & \textbf{HD95} & \textbf{Jaccard(\%)} & \textbf{Precision(\%)} & \textbf{Recall(\%)} & \textbf{Specificity(\%)} \\
\hline
UNet & 31.03M & 66.85$\pm$6.07 & 64.13$\pm$17.53 & 56.80$\pm$5.86 & 63.17$\pm$6.47 & 82.19$\pm$5.34 & 93.99$\pm$3.80 \\
Att-UNet & 57.16M & 62.41$\pm$6.35 & 50.02$\pm$12.15 & 52.90$\pm$6.11 & 56.12$\pm$6.54 & 81.53$\pm$6.02 & 93.98$\pm$3.64 \\
UNet++ & 47.19M & 60.25$\pm$6.56 & 51.33$\pm$15.51 & 51.05$\pm$6.34 & 58.13$\pm$6.56 & 72.24$\pm$7.00 & 82.85$\pm$6.81 \\
TransUNet & 93.23M & 72.54$\pm$5.10 & {\color{blue}49.95$\pm$13.91} & 62.67$\pm$5.39 & {\color{blue}74.36$\pm$5.64} & 77.13$\pm$5.12 & {\color{blue}99.13$\pm$0.17} \\
MISSFormer & 35.45M & 63.67$\pm$5.32 & 79.13$\pm$17.46 & 52.06$\pm$5.17 & 58.60$\pm$5.90 & 79.57$\pm$4.98 & 96.99$\pm$0.69 \\
H-vmunet & 6.44M & 61.89$\pm$5.02 & 71.66$\pm$12.26 & 49.55$\pm$4.86 & 52.86$\pm$5.10 & {\color{blue}90.36$\pm$3.43} & 95.78$\pm$0.84 \\
UltraLight-VM-UNet & 0.037M & 55.32$\pm$5.33 & 66.40$\pm$11.45 & 43.12$\pm$4.97 & 52.38$\pm$5.58 & 69.63$\pm$5.87 & 97.08$\pm$0.56 \\
VM-UNet & 22.04M & {\color{blue}75.16$\pm$5.68} & {\color{red}\textbf{28.33$\pm$8.48}} & {\color{red}\textbf{66.84$\pm$5.46}} & {\color{red}\textbf{78.58$\pm$5.98}} & 74.84$\pm$5.86 & {\color{red}\textbf{99.27$\pm$0.19}} \\
\textbf{Ours} & \textbf{5.65M} & {\color{red}\textbf{75.73$\pm$2.56}} & 53.34$\pm$15.64 & {\color{blue}65.08$\pm$4.70} & 71.46$\pm$4.90 & {\color{red}\textbf{91.16$\pm$2.84}} & 98.28$\pm$0.44 \\
\hline
\end{tabular}%
} 

\vspace{2pt}
\begin{minipage}{\textwidth}
\small
\textit{Notes.} The best results are highlighted in red and the second-best results in blue.
All evaluation metrics are reported as mean $\pm$ standard deviation.
\end{minipage}

\end{table}

\subsection*{4.7 CSAM vs. CBAM Performance Comparison}

Inspired by the classic CBAM~\cite{ref49}, we improve the spatial and channel attention mechanisms, and design the CSAM module within the DAFF framework. Specifically, in the channel attention component, the conventional fully connected layers are replaced with $1 \times 1$ convolutions to reduce the number of parameters. For the spatial attention component, the original $7 \times 7$ convolution is substituted with wavelet convolution to capture richer multi-scale information. The parameter and computational cost statistics of CBAM and CSAM are presented in Table~\ref{tab4}.

\begin{table}[!h]
\caption{Comparison of parameter count and computational cost between the CBAM and CSAM modules.}
\label{tab4}
\centering
\vspace{0.5em}
\begin{tabular}{lcc}
\hline
\textbf{Module} & \textbf{Params} & \textbf{FLOPs} \\
\hline
CBAM  & 111.074K  & 115.848M \\
CSAM  & 41.726K   & 116.236M \\
\hline
\end{tabular}
\end{table}

According to the data in Table~\ref{tab4}, our CSAM module occupies only about 37\% parameters of the CBAM module, while exhibiting nearly identical computational complexity in terms of FLOPs. This indicates that CSAM achieves significantly higher parameter efficiency and lightweight characteristics. Such an advantage makes it particularly suitable for deployment in resource-constrained environments.

\section{Conclusion}
In this paper, the WDFFU-Mamba is proposed to enhance segmentation accuracy for breast tumor in ultrasound images. Building on the Mamba architecture, a Wavelet denoising High-Frequency guided Feature (WHF) module and a Dual Attention Feature Fusion (DAFF) module are developed, significantly improving the model’s capability in boundary representation and feature fusion. Experimental results demonstrate that the proposed method achieves superior performance on both BUSI and BUS, two widely used ultrasound image datasets. In particular, it outperforms existing mainstream models in key metrics such as the Dice coefficient and HD95, indicating stronger boundary awareness and more accurate region overlap. Moreover, despite its relatively low parameter count, the model maintains strong generalization ability, highlighting its potential for deployment in resource-constrained clinical environments. Future work will focus on addressing the challenge of limited annotations in medical imaging, and explore the adaptability and scalability of WDFFU-Mamba in multi-modal scenarios, such as incorporating auxiliary information from textual reports for cross-modal supervision.

\bibliographystyle{unsrt}
\bibliography{cite} 

\begin{thebibliography}{10}

\bibitem{ref1}
Afsaneh Jalalian, Syamsiah~BT Mashohor, Hajjah~Rozi Mahmud, M~Iqbal~B Saripan, Abdul Rahman~B Ramli, and Babak Karasfi.
\newblock Computer-aided detection/diagnosis of breast cancer in mammography and ultrasound: a review.
\newblock {\em Clinical imaging}, 37(3):420--426, 2013.

\bibitem{ref2}
Zhantao Cao, Lixin Duan, Guowu Yang, Ting Yue, Qin Chen, Huazhu Fu, and Yanwu Xu.
\newblock Breast tumor detection in ultrasound images using deep learning.
\newblock In {\em Patch-Based Techniques in Medical Imaging: Third International Workshop, Patch-MI 2017, Held in Conjunction with MICCAI 2017, Quebec City, QC, Canada, September 14, 2017, Proceedings 3}, pages 121--128. Springer, 2017.

\bibitem{ref3}
Olaf Ronneberger, Philipp Fischer, and Thomas Brox.
\newblock U-net: Convolutional networks for biomedical image segmentation.
\newblock In {\em Medical image computing and computer-assisted intervention--MICCAI 2015: 18th international conference, Munich, Germany, October 5-9, 2015, proceedings, part III 18}, pages 234--241. Springer, 2015.

\bibitem{ref4}
Ozan Oktay, Jo~Schlemper, Loic~Le Folgoc, Matthew Lee, Mattias Heinrich, Kazunari Misawa, Kensaku Mori, Steven McDonagh, Nils~Y Hammerla, Bernhard Kainz, et~al.
\newblock Attention u-net: Learning where to look for the pancreas.
\newblock {\em arXiv preprint arXiv:1804.03999}, 2018.

\bibitem{ref5}
Md~Zahangir Alom, Mahmudul Hasan, Chris Yakopcic, Tarek~M Taha, and Vijayan~K Asari.
\newblock Recurrent residual convolutional neural network based on u-net (r2u-net) for medical image segmentation.
\newblock {\em arXiv preprint arXiv:1802.06955}, 2018.

\bibitem{ref6}
Zongwei Zhou, Md~Mahfuzur Rahman~Siddiquee, Nima Tajbakhsh, and Jianming Liang.
\newblock Unet++: A nested u-net architecture for medical image segmentation.
\newblock In {\em Deep learning in medical image analysis and multimodal learning for clinical decision support: 4th international workshop, DLMIA 2018, and 8th international workshop, ML-CDS 2018, held in conjunction with MICCAI 2018, Granada, Spain, September 20, 2018, proceedings 4}, pages 3--11. Springer, 2018.

\bibitem{ref7}
Jieneng Chen, Yongyi Lu, Qihang Yu, Xiangde Luo, Ehsan Adeli, Yan Wang, Le~Lu, Alan~L Yuille, and Yuyin Zhou.
\newblock Transunet: Transformers make strong encoders for medical image segmentation.
\newblock {\em arXiv preprint arXiv:2102.04306}, 2021.

\bibitem{ref8}
Hu~Cao, Yueyue Wang, Joy Chen, Dongsheng Jiang, Xiaopeng Zhang, Qi~Tian, and Manning Wang.
\newblock Swin-unet: Unet-like pure transformer for medical image segmentation.
\newblock In {\em European conference on computer vision}, pages 205--218. Springer, 2022.

\bibitem{ref9}
Yunhe Gao, Mu~Zhou, and Dimitris~N Metaxas.
\newblock Utnet: a hybrid transformer architecture for medical image segmentation.
\newblock In {\em Medical image computing and computer assisted intervention--MICCAI 2021: 24th international conference, Strasbourg, France, September 27--October 1, 2021, proceedings, Part III 24}, pages 61--71. Springer, 2021.

\bibitem{TBME1}
Fang Chen, Haojie Han, Peng Wan, Hongen Liao, Chunrui Liu, and Daoqiang Zhang.
\newblock Joint segmentation and differential diagnosis of thyroid nodule in contrast-enhanced ultrasound images.
\newblock {\em IEEE Transactions on Biomedical Engineering}, 70(9):2722--2732, 2023.

\bibitem{ref10}
Jingkun Yue, Siqi Zhang, Huihuan Xu, Tong Tong, Xiaohong Liu, and Guangyu Wang.
\newblock Ded-sam: Data-efficient distillation towards light-weight medical sam via diverse, balanced, representative core-set selection.
\newblock In {\em 2024 IEEE International Conference on Bioinformatics and Biomedicine (BIBM)}, pages 5418--5425. IEEE, 2024.

\bibitem{ref11}
Jingkun Yue, Siqi Zhang, Zinan Jia, Huihuan Xu, Zongbo Han, Xiaohong Liu, and Guangyu Wang.
\newblock Medsg-bench: A benchmark for medical image sequences grounding.
\newblock {\em arXiv preprint arXiv:2505.11852}, 2025.

\bibitem{ref12}
Jiacheng Ruan, Jincheng Li, and Suncheng Xiang.
\newblock Vm-unet: Vision mamba unet for medical image segmentation.
\newblock {\em arXiv preprint arXiv:2402.02491}, 2024.

\bibitem{ref13}
Gongping Chen, Lu~Zhou, Jianxun Zhang, Xiaotao Yin, Liang Cui, and Yu~Dai.
\newblock Esknet: An enhanced adaptive selection kernel convolution for ultrasound breast tumors segmentation.
\newblock {\em Expert Systems with Applications}, 246:123265, 2024.

\bibitem{ref14}
Jiajia Wang, Guoqi Liu, Dong Liu, and Baofang Chang.
\newblock Mf-net: Multiple-feature extraction network for breast lesion segmentation in ultrasound images.
\newblock {\em Expert Systems with Applications}, 249:123798, 2024.

\bibitem{ref15}
Qiqi He, Qiuju Yang, and Minghao Xie.
\newblock Hctnet: A hybrid cnn-transformer network for breast ultrasound image segmentation.
\newblock {\em Computers in Biology and Medicine}, 155:106629, 2023.

\bibitem{ref16}
Saeid Asgari~Taghanaki, Kumar Abhishek, Joseph~Paul Cohen, Julien Cohen-Adad, and Ghassan Hamarneh.
\newblock Deep semantic segmentation of natural and medical images: a review.
\newblock {\em Artificial intelligence review}, 54:137--178, 2021.

\bibitem{ref17}
Shervin Minaee, Yuri Boykov, Fatih Porikli, Antonio Plaza, Nasser Kehtarnavaz, and Demetri Terzopoulos.
\newblock Image segmentation using deep learning: A survey.
\newblock {\em IEEE transactions on pattern analysis and machine intelligence}, 44(7):3523--3542, 2021.

\bibitem{ref18}
Hanguang Xiao, Zhiqiang Ran, Shingo Mabu, Yuewei Li, and Li~Li.
\newblock Saunet++: an automatic segmentation model of covid-19 lesion from ct slices.
\newblock {\em The Visual Computer}, 39(6):2291--2304, 2023.

\bibitem{ref19}
Bo~Li, Sikai Liu, Fei Wu, GuangHui Li, Meiling Zhong, and Xiaohui Guan.
\newblock Rt-unet: an advanced network based on residual network and transformer for medical image segmentation.
\newblock {\em International Journal of Intelligent Systems}, 37(11):8565--8582, 2022.

\bibitem{ref20}
Shen Jiang and Jinjiang Li.
\newblock Transcunet: Unet cross fused transformer for medical image segmentation.
\newblock {\em Computers in Biology and Medicine}, 150:106207, 2022.

\bibitem{ref21}
Jianyi Zhang, Yong Liu, Qihang Wu, Yongpan Wang, Yuhai Liu, Xianchong Xu, and Bo~Song.
\newblock Swtru: star-shaped window transformer reinforced u-net for medical image segmentation.
\newblock {\em Computers in Biology and Medicine}, 150:105954, 2022.

\bibitem{ref22}
Yang Li, Yue Zhang, Jing-Yu Liu, Kang Wang, Kai Zhang, Gen-Sheng Zhang, Xiao-Feng Liao, and Guang Yang.
\newblock Global transformer and dual local attention network via deep-shallow hierarchical feature fusion for retinal vessel segmentation.
\newblock {\em IEEE Transactions on Cybernetics}, 53(9):5826--5839, 2022.

\bibitem{ref23}
Ashish Vaswani, Noam Shazeer, Niki Parmar, Jakob Uszkoreit, Llion Jones, Aidan~N Gomez, {\L}ukasz Kaiser, and Illia Polosukhin.
\newblock Attention is all you need.
\newblock {\em Advances in neural information processing systems}, 30, 2017.

\bibitem{ref24}
Alexey Dosovitskiy, Lucas Beyer, Alexander Kolesnikov, Dirk Weissenborn, Xiaohua Zhai, Thomas Unterthiner, Mostafa Dehghani, Matthias Minderer, Georg Heigold, Sylvain Gelly, et~al.
\newblock An image is worth 16x16 words: Transformers for image recognition at scale.
\newblock {\em arXiv preprint arXiv:2010.11929}, 2020.

\bibitem{ref25}
Ze~Liu, Yutong Lin, Yue Cao, Han Hu, Yixuan Wei, Zheng Zhang, Stephen Lin, and Baining Guo.
\newblock Swin transformer: Hierarchical vision transformer using shifted windows.
\newblock In {\em Proceedings of the IEEE/CVF international conference on computer vision}, pages 10012--10022, 2021.

\bibitem{ref26}
Ali Hatamizadeh, Vishwesh Nath, Yucheng Tang, Dong Yang, Holger~R Roth, and Daguang Xu.
\newblock Swin unetr: Swin transformers for semantic segmentation of brain tumors in mri images.
\newblock In {\em International MICCAI brainlesion workshop}, pages 272--284. Springer, 2021.

\bibitem{ref27}
Bingzhi Chen, Yishu Liu, Zheng Zhang, Guangming Lu, and Adams Wai~Kin Kong.
\newblock Transattunet: Multi-level attention-guided u-net with transformer for medical image segmentation.
\newblock {\em IEEE Transactions on Emerging Topics in Computational Intelligence}, 8(1):55--68, 2023.

\bibitem{TBME2}
Wenxuan Xu, Cangxin Li, Yun Bian, Qingquan Meng, Weifang Zhu, Fei Shi, Xinjian Chen, Chengwei Shao, and Dehui Xiang.
\newblock Cross-modal consistency for single-modal mr image segmentation.
\newblock {\em IEEE Transactions on Biomedical Engineering}, 2024.

\bibitem{TBME3}
Deepak Mishra, Santanu Chaudhury, Mukul Sarkar, and Arvinder~Singh Soin.
\newblock Ultrasound image segmentation: a deeply supervised network with attention to boundaries.
\newblock {\em IEEE Transactions on Biomedical Engineering}, 66(6):1637--1648, 2018.

\bibitem{ref28}
Huaikun Zhang, Jing Lian, and Yide Ma.
\newblock Fet-unet: Merging cnn and transformer architectures for superior breast ultrasound image segmentation.
\newblock {\em Physica Medica}, 133:104969, 2025.

\bibitem{ref29}
Huaikun Zhang, Jing Lian, Zetong Yi, Ruichao Wu, Xiangyu Lu, Pei Ma, and Yide Ma.
\newblock Hau-net: Hybrid cnn-transformer for breast ultrasound image segmentation.
\newblock {\em Biomedical Signal Processing and Control}, 87:105427, 2024.

\bibitem{ref30}
Huisi Wu, Xiaoting Huang, Xinrong Guo, Zhenkun Wen, and Jing Qin.
\newblock Cross-image dependency modeling for breast ultrasound segmentation.
\newblock {\em IEEE Transactions on Medical Imaging}, 42(6):1619--1631, 2023.

\bibitem{ref31}
Albert Gu and Tri Dao.
\newblock Mamba: Linear-time sequence modeling with selective state spaces.
\newblock {\em arXiv preprint arXiv:2312.00752}, 2023.

\bibitem{ref32}
Lianghui Zhu, Bencheng Liao, Qian Zhang, Xinlong Wang, Wenyu Liu, and Xinggang Wang.
\newblock Vision mamba: Efficient visual representation learning with bidirectional state space model, 2024.

\bibitem{ref33}
Yue Liu, Yunjie Tian, Yuzhong Zhao, Hongtian Yu, Lingxi Xie, Yaowei Wang, Qixiang Ye, Jianbin Jiao, and Yunfan Liu.
\newblock Vmamba: Visual state space model.
\newblock {\em Advances in neural information processing systems}, 37:103031--103063, 2024.

\bibitem{ref34}
Mingya Zhang, Yue Yu, Sun Jin, Limei Gu, Tingsheng Ling, and Xianping Tao.
\newblock Vm-unet-v2: rethinking vision mamba unet for medical image segmentation.
\newblock In {\em International Symposium on Bioinformatics Research and Applications}, pages 335--346. Springer, 2024.

\bibitem{ref35}
Jiashu Xu.
\newblock Hc-mamba: Vision mamba with hybrid convolutional techniques for medical image segmentation.
\newblock {\em arXiv preprint arXiv:2405.05007}, 2024.

\bibitem{ref36}
R~Wu, Y~Liu, P~Liang, and Q~Chang.
\newblock H-vmunet: High-order vision mamba unet for medical image segmentation. arxiv 2024.
\newblock {\em arXiv preprint arXiv:2403.13642}.

\bibitem{ref37}
Chaowei Chen, Li~Yu, Shiquan Min, and Shunfang Wang.
\newblock Msvm-unet: Multi-scale vision mamba unet for medical image segmentation.
\newblock In {\em 2024 IEEE International Conference on Bioinformatics and Biomedicine (BIBM)}, pages 3111--3114. IEEE, 2024.

\bibitem{ref38}
Weibin Liao, Yinghao Zhu, Xinyuan Wang, Chengwei Pan, Yasha Wang, and Liantao Ma.
\newblock Lightm-unet: Mamba assists in lightweight unet for medical image segmentation.
\newblock {\em arXiv preprint arXiv:2403.05246}, 2024.

\bibitem{ref39}
Chao Ma and Ziyang Wang.
\newblock Semi-mamba-unet: Pixel-level contrastive and cross-supervised visual mamba-based unet for semi-supervised medical image segmentation.
\newblock {\em Knowledge-Based Systems}, 300:112203, 2024.

\bibitem{ref40}
Jun Ma, Feifei Li, and Bo~Wang.
\newblock U-mamba: Enhancing long-range dependency for biomedical image segmentation.
\newblock {\em arXiv preprint arXiv:2401.04722}, 2024.

\bibitem{ref41}
Jiarun Liu, Hao Yang, Hong-Yu Zhou, Yan Xi, Lequan Yu, Cheng Li, Yong Liang, Guangming Shi, Yizhou Yu, Shaoting Zhang, et~al.
\newblock Swin-umamba: Mamba-based unet with imagenet-based pretraining.
\newblock In {\em International Conference on Medical Image Computing and Computer-Assisted Intervention}, pages 615--625. Springer, 2024.

\bibitem{ref42}
Zhou Ma, Yunliang Qi, Chunbo Xu, Wei Zhao, Meng Lou, Yiming Wang, and Yide Ma.
\newblock Atfe-net: axial transformer and feature enhancement-based cnn for ultrasound breast mass segmentation.
\newblock {\em Computers in Biology and Medicine}, 153:106533, 2023.

\bibitem{ref43}
Cheng Wang, Le~Wang, Nuoqi Wang, Xiaoling Wei, Ting Feng, Minfeng Wu, Qi~Yao, and Rongjun Zhang.
\newblock Cfatransunet: Channel-wise cross fusion attention and transformer for 2d medical image segmentation.
\newblock {\em Computers in Biology and Medicine}, 168:107803, 2024.

\bibitem{ref44}
Kaiming He, Xiangyu Zhang, Shaoqing Ren, and Jian Sun.
\newblock Deep residual learning for image recognition.
\newblock In {\em Proceedings of the IEEE conference on computer vision and pattern recognition}, pages 770--778, 2016.

\bibitem{ref45}
Walid Al-Dhabyani, Mohammed Gomaa, Hussien Khaled, and Aly Fahmy.
\newblock Dataset of breast ultrasound images.
\newblock {\em Data in brief}, 28:104863, 2020.

\bibitem{ref46}
Moi~Hoon Yap, Gerard Pons, Joan Marti, Sergi Ganau, Melcior Sentis, Reyer Zwiggelaar, Adrian~K Davison, and Robert Marti.
\newblock Automated breast ultrasound lesions detection using convolutional neural networks.
\newblock {\em IEEE journal of biomedical and health informatics}, 22(4):1218--1226, 2017.

\bibitem{ref47}
Xiaohong Huang, Zhifang Deng, Dandan Li, and Xueguang Yuan.
\newblock Missformer: An effective medical image segmentation transformer.
\newblock {\em arXiv preprint arXiv:2109.07162}, 2021.

\bibitem{ref48}
Renkai Wu, Yinghao Liu, Pengchen Liang, and Qing Chang.
\newblock Ultralight vm-unet: Parallel vision mamba significantly reduces parameters for skin lesion segmentation.
\newblock {\em arXiv preprint arXiv:2403.20035}, 2024.

\bibitem{ref49}
Sanghyun Woo, Jongchan Park, Joon-Young Lee, and In~So Kweon.
\newblock Cbam: Convolutional block attention module.
\newblock In {\em Proceedings of the European conference on computer vision (ECCV)}, pages 3--19, 2018.

\end{thebibliography}

\end{document}